\documentclass[letterpaper]{article} 
\usepackage[]{aaai23}  
\usepackage{times}  
\usepackage{helvet}  
\usepackage{courier}  
\usepackage[hyphens]{url}  
\usepackage{graphicx} 
\usepackage{natbib}  
\usepackage{caption} 
\usepackage{algorithm}
\usepackage{algorithmic}

\usepackage{amsfonts,amssymb}
\usepackage{amsmath}
\usepackage{color}
\usepackage[table]{xcolor}
\usepackage{booktabs}
\usepackage{multirow}
\usepackage{makecell}
\usepackage{rotating}
\usepackage{hyperref}

\usepackage[switch]{lineno} 
\usepackage{newfloat}
\usepackage{listings}
\DeclareCaptionStyle{ruled}{labelfont=normalfont,labelsep=colon,strut=off} 
\floatstyle{ruled}
\newfloat{listing}{tb}{lst}{}
\floatname{listing}{Listing}
\title{GLT-T: Global-Local Transformer Voting for 3D Single Object Tracking\\ in Point Clouds}
\author{
    Jiahao Nie\textsuperscript{\rm 1}, Zhiwei He\textsuperscript{\rm 1}\thanks{Corresponding author}, Yuxiang Yang\textsuperscript{\rm 1}, Mingyu Gao\textsuperscript{\rm 1}, Jing Zhang\textsuperscript{\rm 2}
}
\affiliations{
    \textsuperscript{\rm 1 } School of Electronics and Information, Hangzhou Dianzi University, China \\
    \textsuperscript{\rm 2 } School of Computer Science, The University of Sydney, Australia

    $\{$jhnie, zwhe, yyx, mackgao$\}$@hdu.edu.cn, jing.zhang1@sydney.edu.au
}

\usepackage{bibentry}

\begin{document}

\maketitle

\begin{abstract}
Current 3D single object tracking methods are typically based on VoteNet, a 3D region proposal network. Despite the success, using a single seed point feature as the cue for offset learning in VoteNet prevents high-quality 3D proposals from being generated. Moreover, seed points with different importance are treated equally in the voting process, aggravating this defect. To address these issues, we propose a novel global-local transformer voting scheme to provide more informative cues and guide the model pay more attention on potential seed points, promoting the generation of high-quality 3D proposals. Technically, a global-local transformer (GLT) module is employed to integrate object- and patch-aware prior into seed point features to effectively form strong feature representation for geometric positions of the seed points, thus providing more robust and accurate cues for offset learning. Subsequently, a simple yet effective training strategy is designed to train the GLT module. We develop an importance prediction branch to learn the potential importance of the seed points and treat the output weights vector as a training constraint term. By incorporating the above components together, we exhibit a superior tracking method GLT-T. Extensive experiments on challenging KITTI and NuScenes benchmarks demonstrate that GLT-T achieves state-of-the-art performance in the 3D single object tracking task. Besides, further ablation studies show the advantages of the proposed global-local transformer voting scheme over the original VoteNet. Code and models will be available at \url{ https://github.com/haooozi/GLT-T}.
\end{abstract}

\section{Introduction}
3D single object tracking (SOT) in point clouds is a fundamental computer vision task and provides an essential component in various practical applications, such as autonomous driving and mobile robotics \cite{things,spreading}. Given a 3D bounding box (BBox) in the first frame as the target template, the SOT task aims to keep track this target in a sequence of continuous frames \cite{vot2021}.
\begin{figure}[t]
    \centering
    \includegraphics[width=0.95\columnwidth]{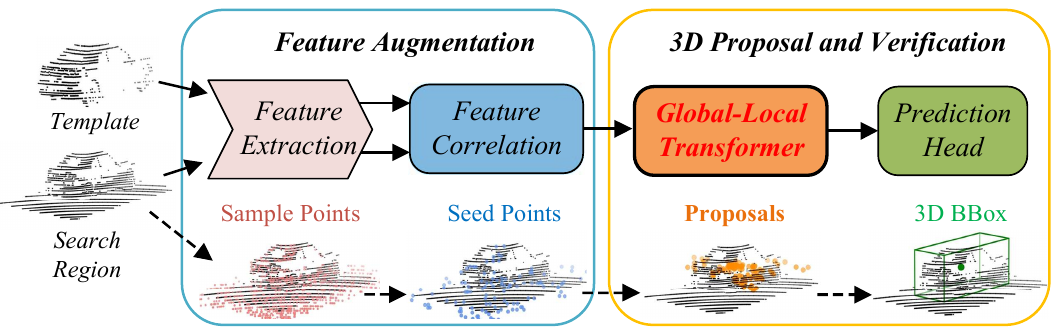} 
    \caption{Schematic illustration to show how \textbf{GTL-T} works. Compared with the existing 3D Siamese tracking methods, we present a global-local transformer voting scheme to generate 3D proposals.}
    \label{fig1}
\end{figure}

Recently, Siamese architecture has achieved great success in 3D SOT. As the pioneer, SC3D \cite{sc3d} uses a Kalman filter to generate a series of candidate 3D BBoxes in current frame, and selects the BBox with the highest similarity to the template as predicted result. However, SC3D is not end-to-end trainable and consumes much computational overhead for matching exhaustive 3D candidate BBoxes. To solve these issues, an end-to-end framework P2B \cite{p2b} is proposed. It introduces a 3D region proposal network (RPN), i.e., VoteNet \cite{votenet} to generate 3D proposals and predict the best one as tracking BBox. Benefiting from the VoteNet, P2B obtains excellent performance in terms of both accuracy and efficiency. Nevertheless, some inherent defects in VoteNet prevent high-quality proposals from being generated, thereby limiting the tracking performance. Different from the follow-up works \cite{ptt, mlvsnet} that have rarely investigated the voting scheme of VoteNet, we point out its two defects as follows:
\begin{itemize}
    \item The voting scheme uses only a single point feature as the cue to infer the offsets of seed points to the target centers, making it difficult to generate high-quality proposals due to the limited representation ability of point feature.
    
    \item Seed points located at different geometric positions in the target objects contribute differently to learning the offsets. However, existing methods treat them equally in the voting scheme, greatly distracting the model in training and leading to sub-optimal proposals.
\end{itemize}

In this paper, we propose a novel \textbf{g}lobal-\textbf{l}ocal \textbf{t}ransformer voting scheme for \textbf{t}racking (GLT-T) to address the above challenges. A schematic illustration of GLT-T is shown in Fig. \ref{fig1}. To provide more informative cues for offset learning, we revisit the feature representation requirement in voting process and suggest a reasonable assumption: modeling feature prior about the geometric position of seed points in the target object can significantly assist in learning the offsets of seed points to the target centers. To achieve this goal, we propose to integrate object- and patch-aware prior into the voting network to infer the position of seed points, providing more robust and accurate cues for offset learning. Motivated by that point transformer \cite{pt} has powerful global and local modeling capabilities as well as its inherent permutation invariance, we elaborately design the global-local transformer (GLT) module, a cascaded network for voting. In contrast to the original point transformer that only encodes the intensity features of point clouds, the proposed cascaded transformer structure also effectively encodes the geometric features. Thus, the GLT module can model the useful geometric position prior of seed points located in the target object, guiding more accurate offset learning. Specifically, global-level self-attention and local-level self-attention are designed as the core components. On the one hand, the global-level self-attention encodes the geometric feature of each seed point by integrating global object information. On the other hand, the local-level self-attention enhances the geometric feature representation of each seed point by integrating local patch information.

To emphasize the different importance of seed points, making the voting network pay more attention on the seed points that enjoy large potential to offset to the target centers. We therefore propose to address this challenge by introducing a weights vector as the constraint term to train the GLT module. The weights vector plays an important role in emphasizing seed points, since it describes how difficult the seed points offset to the target centers. Inspired by the concept of centerness \cite{fcos} used in 2D object detection \cite{detection1, detection2}, which represent the potential object centers, we propose an importance prediction branch to learn this weight vector. Generally, the seed points with rich object location information are prone to offset to the target centers, and deserve much attention. Thus, we attach this branch to the global-transformer block that encodes the object-aware geometric prior to learn the importance of seed points.

Our main contributions can be summarized as follows:
\begin{itemize}
    \item We propose a novel GLT-T method for 3D single object tracking, alleviating the long-standing unresolved voting problem by a GLT module and a training strategy. To the best of our knowledge, this is the first attempt to improve the voting scheme.
    \item GLT module leverages both the global-level and local-level self-attention to encode object- and patch-aware prior and provide informative cues for offset learning, thus producing high-quality 3D proposals and significantly enhancing tracking performance.
    \item Extensive experiments on KITTI \cite{kitti} and NuScenes \cite{nuscenes} show that GLT-T achieves SOTA performance.
\end{itemize}

\section{Related Work}
\subsection{3D Single Object Tracking}
Early 3D SOT methods \cite{rgbd1, rgbd2, rgbd3, rgbd4, rgbd5} mainly track objects in RGB-D domain. Although delivering promising results, the RGB-D trackers may fail to track the targets when the RGB-D information is degraded, e.g., due to illumination and weather variations. Since 3D point cloud data captured by LiDAR sensors is insensitive to the these variations, point cloud-based SOT has attracted great attention recently. SC3D \cite{sc3d} is the first 3D Siamese tracker using pure point clouds. P2B \cite{p2b} introduces a 3D region proposal network (RPN) (i.e., VoteNet \cite{votenet}) and achieves SOTA performance in terms of both accuracy and efficiency. It adopts a tracking-by-detection scheme, which employs a VoteNet to generate 3D proposals and predicts the proposal with the highest score for tracking. Inspired by this strong baseline, many follow-up works have been proposed. BAT \cite{bat} propose a box-aware feature enhancement module to replace the point-wise correlation operation in P2B. Besides, LTTR \cite{lttr} and PTTR \cite{pttr} also propose different correlation operations using 3D transformers to promote feature interactions. However, due to the aforementioned issues of the common component VoteNet, the tracking performance is limited by the unsatisfactory proposals.
\begin{figure*}[t]
    \centering
    \includegraphics[width=1.9\columnwidth]{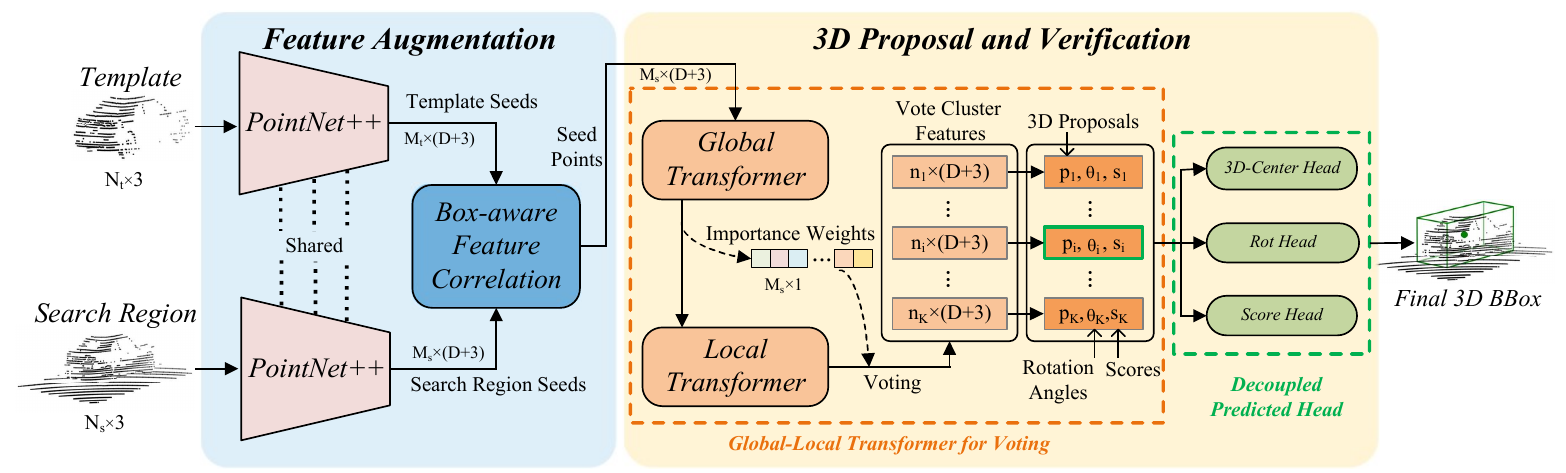} 
    \caption{Overview of our proposed \textbf{GLT-T}. Given a template and search region, we first utilize PointNet++ \cite{pointnet++} to extract the point features, and fuse them with a box-aware feature correlation \cite{bat} to output seed points. We then apply the proposed global-local transformer voting scheme to generate 3D proposals. Finally, we perform decoupled prediction head to verify the input proposals and output a final 3D BBox.}
    \label{fig2}
\end{figure*}

To solve this limitation, MLVSNet \cite{mlvsnet} presents a multi-level voting strategy that uses the multi-layer features of backbone for voting. Similarly, PTT \cite{ptt} proposes a feature enhancement module with transformer to guide powerful seed point features that participate in voting. However, they fail to consider improving the feature representation for offset learning, and give equal attention to seed points. In this paper, we deeply investigate the voting procedure and the learning of better feature representation. To this end, we develop a novel voting scheme with a GLT module and a training strategy to guide more accurate offset learning and produce high-quality 3D proposals, thus improving the tracking performance.

\subsection{Transformer}
Transformer is originally proposed in the area of natural
language processing \cite{aiayn}, showing an excellent ability in modeling long-range dependency. It becomes popular in computer vision recently and has been widely used for image classification \cite{liu2021swin,xu2021vitae,zhang2022vitaev2,zhang2022vsa}, object detection \cite{detr,wang2021fp,wang2022towards}, pose estimation \cite{hpe}, and object tracking \cite{lan,learning}. The success motivates researchers to extend it to 3D point cloud tasks. Due to the operations such as linear layer, softmax and normalization maintain the permutation invariance of point clouds, 3D transformers are well suited for point cloud processing \cite{pt, pct,ptsurvey}.

In 3D SOT, several transformer tracking methods \cite{lttr,pttr,ptt} have been proposed. They either use self-attention to further process features or use cross-attention to interact features from template and search region. In contrast, we elaborate the global-level and local-level self-attention in GLT module to encode object- and patch-aware prior, allowing to provide informative cues for offset learning. Moreover, we emphasize the more important seed points by introducing a novel importance prediction branch.

\section{Methodology}
\subsection{Overview}
For 3D object tracking, a template point cloud $P^{t}=\{p_i^t\}_{i=1}^{N_t}$ together with the 3D BBox $B_t=(x_t,y_t,z_t,w_t,h_t,l_t,\theta_t)$ in initial frame are given, where $(x,y,z)$ and $(w,h,l)$ denotes the center coordinate and size, $\theta$ is the rotation angle around $up$-axis. The goal is to locate this template target in search region $P^{s}=\{p_i^s\}_{i=1}^{N_s}$ and output a 3D BBox $B_s=(x_s,y_s,z_s,\theta_s)$ frame by frame, where $N_t$ and $N_s$ denote the number of input points for template and search region. Notably, since the target size is kept constant in all frames, we output only 4 parameters to represent $B_s$. The tracking process can be formulated as:
\begin{equation}
    track: {\text{GLT-T}}(P^t,P^s)\to (x_s,y_s,z_s,\theta_s),
    \label{eq1}
\end{equation}
where GLT-T is detailed in Fig. \ref{fig2}. It involves two stages to perform tracking: 1) Feature Augmentation, and 2) 3D Proposal and Verification. In the first stage, we employ PointNet++ \cite{pointnet++} as the Siamese backbone to extract point features and output the seed points by a box-aware feature correlation presented in BAT \cite{bat}. In the second stage, we propose a global-local-transformer voting method to generate 3D proposals $\{(x_p,y_p,z_p)\}_{p=1}^K$, where $K$ denotes the number of proposals. Furthermore, we decouple the prediction head to predict the confidence scores of the 3D proposals and their rotation angles $\{\theta_p\}_{p=1}^K$. The proposal $(x_p,y_p,z_p)$ with the highest confidence score and the corresponding $\theta_p$ jointly represent the final 3D BBox.

\subsection{Global-Local Transformer for Voting}
\begin{figure}[t]
    \centering
    \includegraphics[width=0.9\columnwidth]{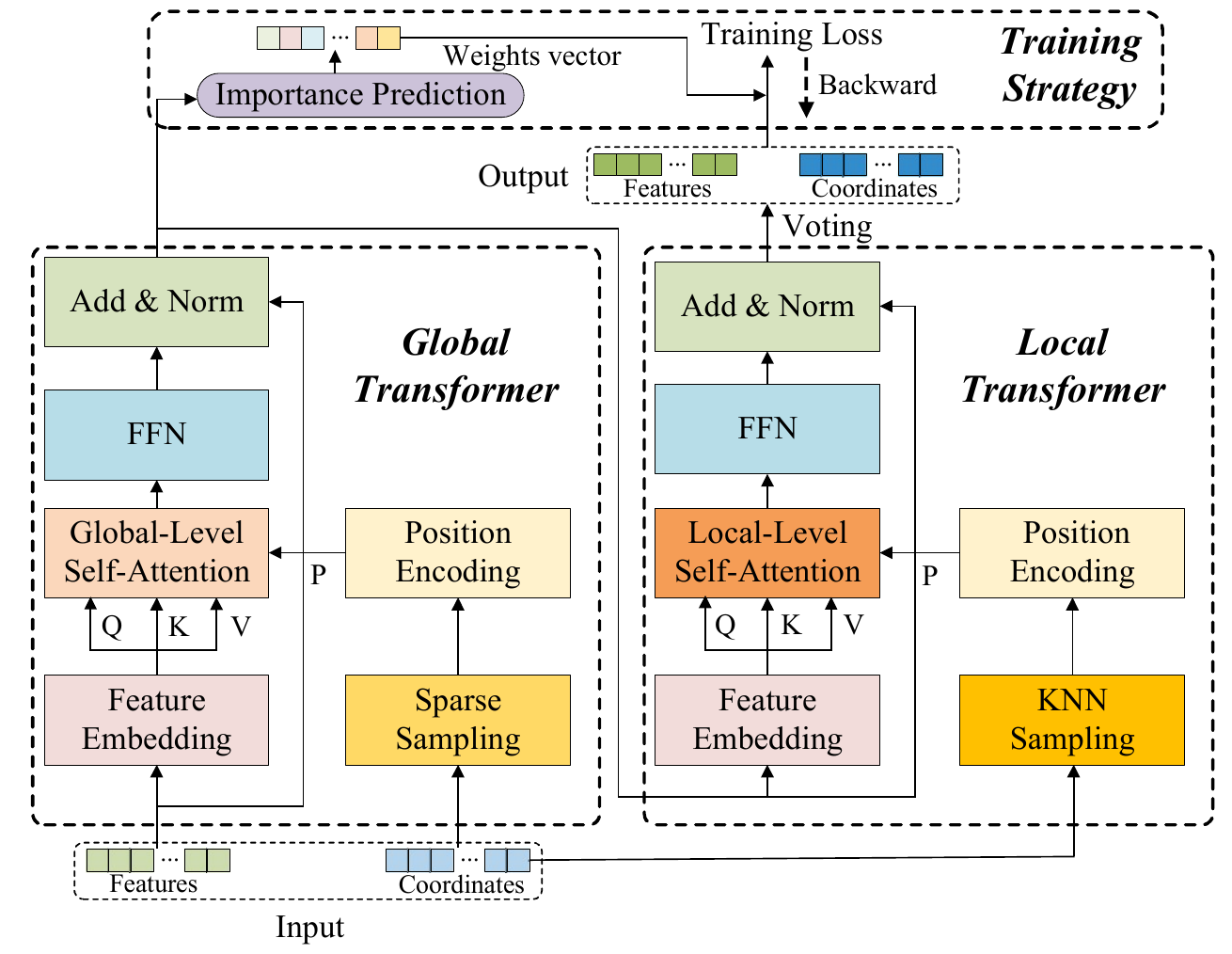}
    \caption{Schematic illustration of GLT voting scheme. It consists of three components: a Global Transformer, a Local Transformer and a Training Strategy. Given the seed points including features and coordinates as inputs, the cascaded global-local transformer encode the object- and patch-aware information for seed points to perform voting. An importance prediction branch is bridged after the global transformer block, which is used to learn importance weights of seed points to constrain the training process.}
    \label{fig3}
\end{figure}

We devise a global-local transformer based voting scheme to learn accurate offsets. Specifically, the seed points located on the target surface are used to regress the target center and generate accurate 3D proposals $\{(x_p,y_p,z_p)\}_{p=1}^K$. As shown in Fig. \ref{fig3}, our voting scheme has three essential components: a Global Transformer, a Local Transformer and a Training Strategy. Given the seed points $S=\{s_i\}_{i=1}^{M_s}$, where $s_i=[f_i;c_i]\in \mathbb{R}^{D+3}$, $f_i\in\mathbb{R}^D$ and $c_i\in\mathbb{R}^3$ represent the $D$-dimension features and 3D coordinate of point $s_i$. We first use the global transformer block to encode the object-aware information for the seed points: $[f_i;c_i]\to f_i^g$. Afterwards, the local transformer block is used to further encode the patch-aware information: $[f_i^g;c_i]\to f_i^{gl}$. With the cascaded global-local transformer module, informative cues are integrated into the seed points for voting. In addition, an importance prediction branch induced by the global transformer block is introduced to learn the weights vector, which assigns importance weights to the seed points in training.

\begin{figure}[t]
    \centering
    \includegraphics[width=0.6\columnwidth]{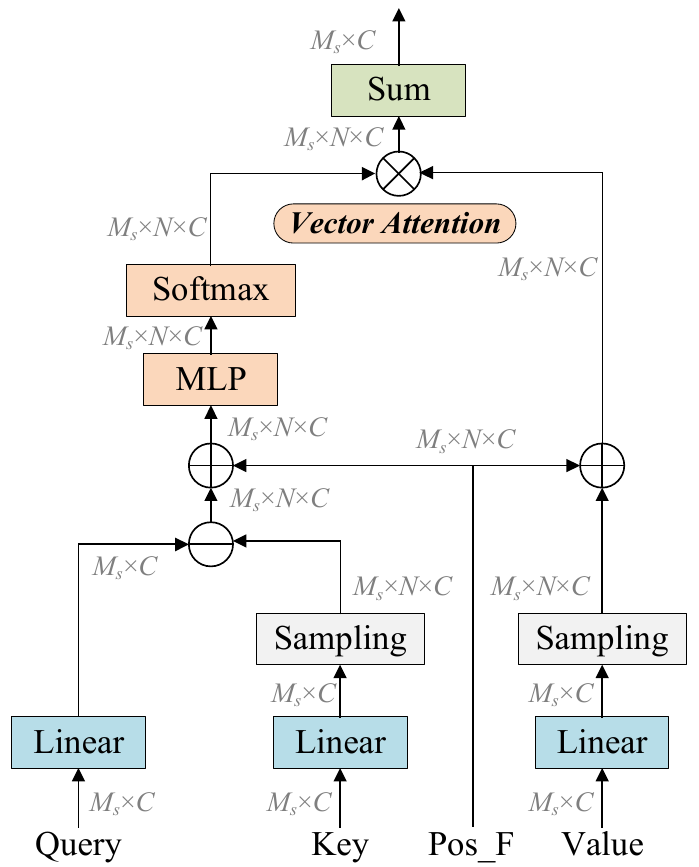} 
    \caption{Illustration of the vector attention. It first projects query, key and value into a latent space and samples the global / local region features from the key and value ($N=m$ and $N=n$ for global-level self-attention and local-level self-attention, respectively).Then a MLP and a Softmax operator are employed to generate the attention matrix. Based on it, we apply a product dot on the value and obtain the final output by calculating the sum along the attention dimension.}
    \label{fig4}
\end{figure}

\noindent\textbf{Global Transformer.} The Global Transformer block takes the seed points  $S=\{s_i\}_{i=1}^{M_s}$ as its inputs. As shown in Fig. \ref{fig3}, we use a linear layer to embed the original features $\{f_i\}_{i=1}^{M_s}$ into $\{f_i^*\}_{i=1}^{M_s}$ to obtain query (Q), key (K) and value (V), where $f_i^*\in \mathbb{R}^C$. 
Meanwhile, position encoding is added. It enables the attention operation to adapt to different ranges of information. To model object-aware prior and account for computational efficiency, we propose a sparse sampling strategy to provide global geometric shape information $c_{iss}$ for each seed point $c_i$. The success of this sparse sampling is owing to the feature similarity of neighboring points.
In practice, we first calculate the distance matrix $\{D_{i,j}\}_{i=1,j=1}^{M_s,M_s}$ of the seed points by:
\begin{equation}
    D_{i,j}=\parallel c_i-c_j\parallel_2, \forall i,j\in\{1,2,...,M_s\},
    \label{eq2}
\end{equation}
where $\parallel\cdot\parallel_2$ denotes L2-norm. Then, each row of the matrix $\{D_{i,j}\}_{i=1,j=1}^{M_s,M_s}$ is ordered from smallest to largest, and $m$ seed points are sparsely sampled as the $c_{iss}$ as follows: 
\begin{equation}
    c_{iss}=\mathop{{\rm sort}} \limits_{j=1}^{M_s}(D_{i,j})[::\frac{M_s}{m}], \forall i\in\{1,2,...,M_s\}.
    \label{eq3}
\end{equation}
Given the global geometric shape information $c_{iss}\in \mathbb{R}^{m\times 3}$, we consider using the relative position between $c_i$ and $c_{iss}$, i.e., $c_i-c_{iss}$, which captures geometric spatial information, to get the position encodings $Pos\_F=\{pos\_f_i\}_{i=1}^{M_s}$:
\begin{equation}
    pos\_f_i=\varphi_g(c_i-c_{iss}),
    \label{eq4}
\end{equation}
where $pos\_f_i \in \mathbb{R}^{m \times C}$ and $\varphi_g(\cdot)$ is a two-layer MLP network with ReLU \cite{relu}.

After getting the feature embedding and position encoding, the global-level self-attention is designed to integrate the features of global sparse points for each seed point. We use a vector attention operator (as shown in Fig. \ref{fig4}) to perform information integration since it can model both channel and spatial information interactions, which is more efficient than scalar attention for point cloud transformer. Mathematically, the global-level self-attention can be formulated as:
\begin{equation} 
    \begin{aligned}
        \{f_i^{glsa}\}_{i=1}^{M_s} & = {\rm GLSA}(Q,K,V,Pos\_F) \\
        & = {\rm Softmax}(\phi(\omega_q(Q)-\omega_{ks}(K)\\
        & \quad +Pos\_F)) \cdot(\omega_{vs}(V)+Pos\_F),
    \end{aligned}
    \label{eq5}
\end{equation}
where $\omega_q(\cdot)$ is a linear layer, $\omega_{ks}/\omega_{vs}(\cdot)$ is a linear layer with sparse sampling and  $\phi(\cdot)$ is a two-layer MLP network with ReLU. Finally, $\{f_i^{glsa}\}_{i=1}^{M_s}$ is fed into a feed-forward network, and then added with the original features $\{f_i\}_{i=1}^{M_s}$ to get the final output $\{f_i^g\}_{i=1}^{M_s}$:
\begin{equation}
    \{f_i^g\}_{i=1}^{M_s}={\rm Norm}({\rm FFN}(\{f_i^{glsa}\}_{i=1}^{M_s})+\{f_i\}_{i=1}^{M_s}),
    \label{eq6}
\end{equation}
where FFN is a linear layer that maps the size $\mathbb{R}^C$ to $\mathbb{R}^D$, and Norm denotes a layer normalization function to increase the fitting ability of the network.

\noindent\textbf{Local Transformer.} To further provide powerful cues for voting, we use the Local Transformer block to model patch-aware prior to enhance the geometric feature representation. The local transformer block has a similar structure to the global transformer block as shown in Fig. \ref{fig3}. Differently, we extract the local region information $c_{iks}$ around $c_i$ by K-nearest-neighbor (KNN) sampling:
\begin{equation}
    c_{iks}=\mathop{{\rm min}} \limits_{j=1}^{n}(D_{i,j}), \forall i\in\{1,2,...,M_s\},
    \label{eq7}
\end{equation}
where $c_{iks}\in \mathbb{R}^{n\times 3}$, $n$ controls the size of the local region. By encoding the relative position between each seed point and the corresponding $n$ nearest neighbor points, the geometric feature representation of each seed point is enhanced, which is beneficial to learn offsets to the target center.

With the enhanced seed point features $\{f_i^{gl}\}_{i=1}^{M_s}$ that are incorporated with both object- and patch-aware prior, a voting module is employed to learn the offsets:
\begin{equation}
    \begin{aligned}
     [\Delta f_i^v; \Delta c_i^v] & = {\rm Voting}([f_i^{gl};c_i]) \\ 
     [f_i^v;c_i^v] & =[f_i^{gl};c_i] + [\Delta f_i^v; \Delta c_i^v],
     \end{aligned}  
    \label{eq8}
\end{equation}
where Voting($\cdot$) is a three-layer MLP network with batch normalization and ReLU. It learns the feature residuals and coordinate offsets for the seed points participating in the voting process. At the end, we sample $K$ ($K<M_s$) 3D proposals $\{(x_p,y_p,z_p)\}_{p=1}^K$ from $\{c_i^v\}_{i=1}^{M_s}$ using farther point sampling (FPS) \cite{pointnet}.

\noindent\textbf{Training Strategy.} Considering that the seed points have different impact on learning the offset to the target center, as well as only one proposal is predicted as final 3D BBox, we propose a training strategy to distinguish the importance of seed points in the voting process, enabling the network to focus on the seed points that are easier to offset to the target center. To this end, an importance prediction branch bridged behind the global transformer block is introduced to learn the weights vector, as shown in Fig. \ref{fig3}. This branch is implemented as a three-layer MLP, and the weights vector $\{I_i\}_{i=1}^{M_s}$ is calculated by:
\begin{equation}
    I_i={\rm MLP}(f_i^g),
    \label{eq9}
\end{equation}
where $I_i$ denotes the importance weight of the $i$-th seed point. To make this weights vector able to represent the importance of seed points, we use supervised learning to train this branch. The seed points within the target object are defined as positive samples and the others are defined as negative samples. Then we use a binary cross entropy loss $L_{bce}$ for $\{I_i\}_{i=1}^{M_s}$, and the optimization objective is defined as:
\begin{equation}
    \mathcal{L}_{imp}=\frac{1}{M_s}\sum_{i=1}^{M_s}L_{bce}(I_i,o_i),
    \label{eq10}
\end{equation}
where $o_i$ denotes the label for $i$-th seed point, i.e., $o_i=1$ and $o_i=0$ for positive and negative sample, respectively. Using this weights vector, we assign importance weights for seed points in training. In fact, higher weighted points deserve more attention, so we use a modified smooth L1 loss for offset learning:
\begin{equation}
    \mathcal{L}_{off}=\frac{1}{\sum_{i=0}^{M_s}{o_i}}\sum_{i=0}^{M_s}{\rm smooth_{L1}}(c_i^v,c^l)\cdot(1+I_i)\cdot o_i,
    \label{eq11}
\end{equation}
where $c[0]=x,c[1]=y,c[2]=z$ and $c^l$ is the ground truth of target center. As illustrated in Eq. \ref{eq11}, the loss for different seed points is dynamically adjusted according to $\{I_i\}_{i=1}^{M_s}$.

\subsection{Decoupled Prediction Head}
Most existing trackers use a single MLP as the coupled prediction head, which simultaneously predicts the scores, rotation angles, and refined locations of the proposals. However, the three sub-tasks require different features, so the coupled prediction head leads to suboptimal performance. To solve this issue, we decouple the prediction head and use three MLP networks to perform the prediction task, in which each MLP corresponds to one sub-task. Specifically, as shown in Fig. \ref{fig2}, the decoupled prediction head consists of three sub-prediction heads: 3d-center head, rotation angle head and score head, and each prediction head is composed of a three-layer MLP network with normalization and ReLU. The first two linear layers keep the feature dimension fixed, while the last layer aligns the output feature dimension according to the different sub-tasks. With a slight increase in model complexity, the tracking performance is effectively improved by the decoupled prediction head.

\subsection{Training Loss} The overall loss arises from two sources, i.e., voting network and prediction head. It is defined as:
\begin{equation}
    \mathcal{L}=\mathcal{L}_{off}+\lambda_1\mathcal{L}_{imp}+\lambda_2\mathcal{L}_{score}+\lambda_3\mathcal{L}_{center,rot},
    \label{eq12}
\end{equation}
where $\lambda_1$, $\lambda_2$ and $\lambda_3$ are hyper-parameters to balance different losses. 
Details about $\mathcal{L}_{score}$ and $\mathcal{L}_{center,rot}$ can be found in \cite{p2b}.

\section{Experiments}
\begin{table*}[!t]
    \centering
    \begin{tabular}{cccccc}
      \toprule
      Category & Car & Pedestrian &  Van & Cyclist & Mean\\
      Frame Number & 6424 & 6088 & 1248 & 308 & 14068 \\
      \midrule
      \midrule
      SC3D \cite{sc3d}& 41.3 / 57.9 & 18.2 / 37.8 & 40.4 / 47.0 & 41.5 / 70.4 &31.1 / 48.5 \\
      P2B \cite{p2b}& 56.2 / 72.8 & 28.7 / 49.6 & 40.8 / 48.4 & 32.1 / 44.7 &42.4 / 60.0 \\
      F-Siamese \cite{f-siamese}& 37.1 / 50.6 & 16.2 / 32.2 & - / - & 47.0 / 77.2 &- / - \\
      3D-SiamRPN \cite{3d-siamrpn}& 58.2 / 76.2 & 35.2 / 56.2 & 45.6 / 52.8 & 36.1 / 49.0 &46.6 / 64.9 \\
      PTT \cite{ptt}& 67.8 / \underline{81.8} & 44.9 / 72.0 & 43.6 / 52.5 & 37.2 / 47.3 & 55.1 / 74.2 \\
      LTTR \cite{lttr}& 65.0 / 77.1 & 33.2 / 56.8 & 35.8 / 48.4 & \underline{66.2} / 89.9 &48.7 / 65.8 \\
      MLVSNet \cite{mlvsnet}& 56.0 / 74.0 & 34.1 / 61.1 & 52.0 / 61.4 & 34.3 / 44.5 & 45.7 / 66.6 \\
      BAT \cite{bat}& 60.7 / 74.9 & 42.1 / 70.1 & 31.5 / 38.9 & 53.0 / 82.5 & 50.0 / 69.9 \\
      V2B \cite{v2b}& \textbf{70.5} / 81.3 & 48.3 / 73.5 & 50.1 / 58.0 & 40.8 / 49.7 &\underline{58.4} / 75.2 \\
      PTTR \cite{pttr}& 65.2 / 77.4 & \underline{50.9} / \textbf{81.6} & \underline{52.5} / \underline{61.8} & 65.1 / \underline{90.5} &57.9 / \underline{78.2} \\
      \midrule
      \midrule
      GLT-T (ours) & \underline{68.2} / \textbf{82.1} & \textbf{52.4} / \underline{78.8} & \textbf{52.6} / \textbf{62.9} & \textbf{68.9} / \textbf{92.1} &  \textbf{60.1} / \textbf{79.3} \\
      \bottomrule
    \end{tabular}
    \caption{Performance comparison with SOTA methods on the KITTI dataset. \textit{Success} / \textit{Precision} are used for evaluation. \textbf{Bold} and \underline{underline} denote the best result and the second-best one, respectively}
    \label{table1}
\end{table*}

\begin{table*}[!ht]
    \centering
    \begin{tabular}{cccccc}
      \toprule
      Category & Truck & Trailer &  Bus & Motorcycle & Mean\\
      Frame Number & 13587 & 3352 & 2953 & 2419 & 22311 \\
      \midrule
      \midrule
      SC3D \cite{sc3d}& 30.7 / 27.7 & 35.3 / 28.1 & 29.4 / 24.1 & 17.2 / 24.8 & 29.8 / 27.0 \\
      P2B \cite{p2b}& 42.9 / 41.6 & 48.9 / 40.1 & 32.9 / 27.4 & 21.3 / 33.4 &40.1 / 38.6 \\
      BAT \cite{bat}& \underline{45.3} / \underline{42.6} & \underline{52.6} / \underline{44.9} & \underline{35.4} / \underline{28.0} & \underline{22.7} / \underline{35.6} & \underline{42.6} / \underline{40.1} \\
      \midrule
      \midrule
      GLT-T (ours) & \textbf{52.7} / \textbf{51.4} & \textbf{57.6} / \textbf{52.0} & \textbf{44.6} / \textbf{40.7} & \textbf{34.8} / \textbf{47.6} &\textbf{50.4} / \textbf{49.7} \\
      \bottomrule
    \end{tabular}
    \caption{Performance comparison with SOTA methods on the NuScenes dataset. \textit{Success} / \textit{Precision} are used for evaluation. \textbf{Bold} and \underline{underline} denote the best result and the second-best one, respectively}
     \label{table2}
\end{table*}

\begin{figure*}[t]
    \centering
    \includegraphics[width=1.95\columnwidth]{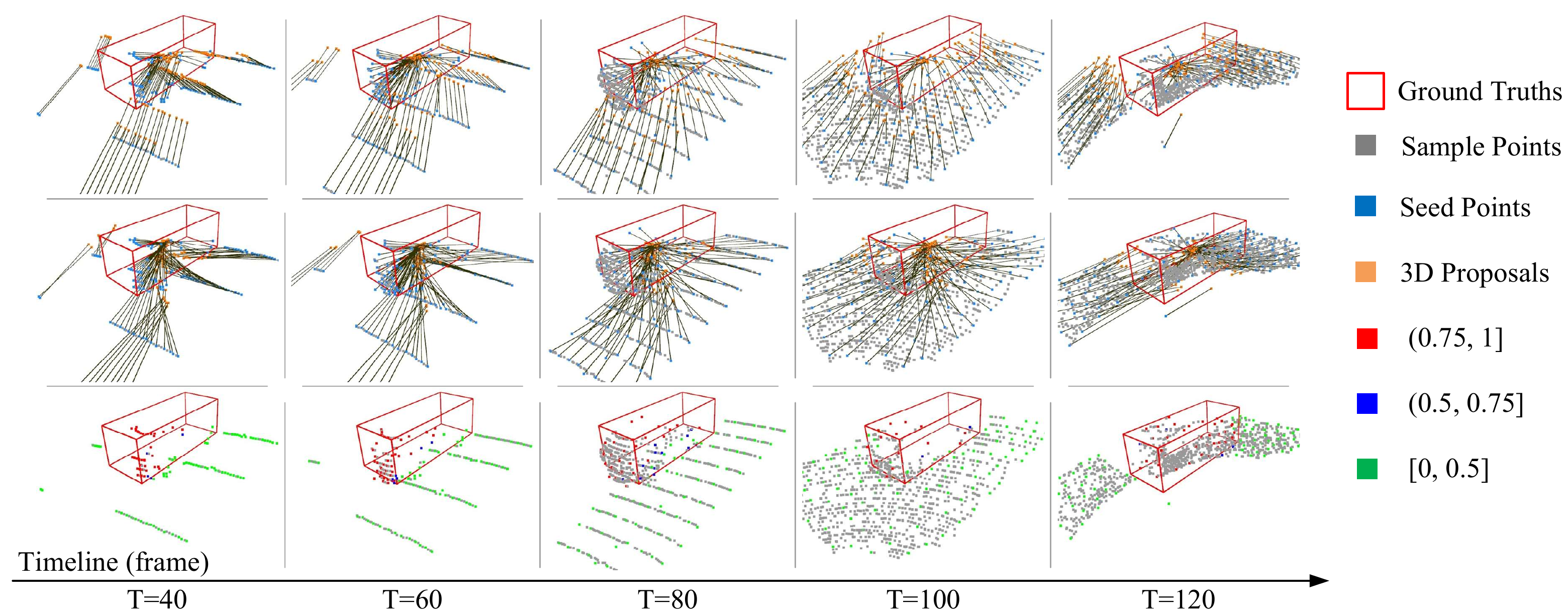} 
    \caption{Visualization the voting process of VoteNet \cite{votenet} (top row) and our voting scheme (middle row) on a point cloud sequence from KITTI. The red boxes denote the ground truth. The sample points are colored by grey, while the seed points are colored by blue and the 3D proposals regressed by seed points are colored by orange. In the bottom row, we show the predicted importance scores of seed points in red, blue and green for the ranges of (0.75,1], (0.5,0.75] and [0,0.5], respectively.}
    \label{fig5}
\end{figure*}

\subsection{Experiment Setup}
\noindent\textbf{Datasets.} Following previous works, KITTI \cite{kitti} and NuScenes \cite{nuscenes} datasets are adopted to train and test the proposed GLT-T method. KITTI contains 21 training LiDAR sequences and 29 test LiDAR sequences, which are sampled at 10Hz. Since the test sequence is not open, we follow the common protocol and divide the training LiDAR sequences into training set (0-16), validation set (17-18) and test set (19-20). NuScenes is a more challenging dataset, containing 1,000 scenes and providing the LiDAR annotations at 2Hz. Following BAT \cite{bat}, we train our GLT-T model on training set, and evaluate the performance on validation set with key frames.

\noindent\textbf{Evaluation Metrics.} Following the common practice, we use One Pass Evaluation (OPE) \cite{otb} to measure the metrics \textit{Success} and \textit{Precision} and report the performance of different trackers. Given a predicted BBox and the corresponding ground truth BBox, the Intersection Over Union (IoU) between them is defined as overlap and the distance between their centers is defined as error. \textit{Success} denotes the Area Under the Curve (AUC) with the overlap threshold ranging from 0 to 1, while \textit{Precision} denotes the AUC with the error threshold ranging from 0 to 2 meters.

\subsection{Comparison with SOTA methods}
\noindent\textbf{Results on KITTI.} We compare our GLT-T with all relevant SOTA methods. The results of four categories, including Car, Pedestrian, Van and Cyclist, as well as their mean results are presented in Table \ref{table1}. As reported, GLT-T outperforms all comparison methods in overall performance. Specifically, we obtain the best \textit{Success} or \textit{Precision} values under all categories. In particular, for Van and Cyclist categories with small instances, GLT-T achieves superior performance, implying that the GLT module can still effectively form strong object- and patch-aware prior under the condition of small training samples. Furthermore, compared to the SOTA method BAT, GLT-T exhibits a significant performance gain by 10.1\% and 9.4\% in terms of mean \textit{Success} and \textit{Precision}, respectively. Although BAT achieves the impressive performance in Car and Pedestrian categories, our method GLT-T still outperforms it by a remarkable advantage. The improvements are attributed to our voting scheme, which can generate high-quality 3D proposals and thus improves the tracking accuracy. The voting scheme is further analyzed in the follow-up ablation study.

\noindent\textbf{Results on NuScenes.} To further evaluate GLT-T, comparative experiments are conducted on the more challenging dataset NuScenes with four different categories, including Truck, Trailer, Bus and Motorcycle. We select the trackers that have reported performance on the Nuscenes dataset as comparisons. As shown in Table \ref{table2}, GLT-T achieves the best performance and outperforms BAT by 7.8\% (\textit{Success}) and 9.6\% (\textit{Precision}). Besides, due to the various outdoor scenes involved in this dataset, the promising performance of our method proves its generalizability to real-world tracking point cloud sequences.

\subsection{Ablation Study}
\noindent\textbf{Model Components.} In GLT-T, we propose a novel voting scheme to generate 3D proposals. It consists of two key component: GLT module and training strategy. To verify the effectiveness of the components, a comprehensive ablation experiment is conducted on the Car category of KITTI dataset following \cite{p2b,bat}. We report the results in Table \ref{table3}. By using the GLT module, our method achieve a performance improvement of 4.7\% and 4.6\% in terms of \textit{Success} and \textit{Precision}, respectively. When using the proposed training strategy to train this module, the performance improvement increases up to 6.5\% and 6.0\%.

\begin{table}[!t]
    \centering
    \begin{tabular}{ccc}
      \toprule
      Tracking Variations & \textit{Success} & \textit{Precision}\\
      \midrule
      \midrule
      w/o GLT, w/o TS &  61.7 & 76.1 \\
      w/ GLT, w/o TS & 66.4$_{\uparrow4.7}$  & 80.7$_{\uparrow4.6}$ \\
      GLT-T: w/ GLT, w/ TS & 68.2$_{\uparrow6.5}$  & 82.1$_{\uparrow6.0}$ \\
      \bottomrule
    \end{tabular}
    \caption{Ablation study on the proposed voting scheme, including the GLT module and training strategy (TS).}
    \label{table3}
\end{table}

To intuitively show the superiority of our voting scheme over the VoteNet \cite{votenet}, we visualize their voting process in Fig. \ref{fig5}. It can be clearly observed that our approach generates more 3D proposals that are close to the target centers. This is owing to the fact that the GLT module integrates both object- and part-aware information for offset learning. Besides, due to the importance scores for seed points in our voting scheme, the seed points within a certain region have diverse offset directions and values, making the offsets of seed points with different geometric positions to the target centers more accurate. By contrast, the neighboring seed points in VoteNet present similar offsets. In the bottom row, we show the predicted importance scores. The seed points in target objects have higher scores, indicating that the training strategy emphasizes the more important seed points while effectively discriminating those far-away ones.

\noindent\textbf{GLT Module.} We further investigate the effectiveness of each transformer block, and the ablation results are shown in Table \ref{table4}. The global transformer block encodes the geometric position features of the seed points by integrating object-aware prior, which allows to infer accurate offsets to generate 3D proposals. Therefore, 3.1\% (\textit{Success}) and 2.9\% (\textit{Precision}) performance improvements are achieved. Besides, since the local transformer integrates parch-aware prior into each seed point, enhancing the feature representation of the geometric position. The tracking performance is further improved. 

\begin{table}[!ht]
    \centering
    \begin{tabular}{ccc}
      \toprule
      Tracking Variations & \textit{Success} & \textit{Precision} \\
      \midrule
      \midrule
      w/o GT, w/o LT &  61.7 & 76.1 \\
      w/ GT, w/o LT & 64.8$_{\uparrow3.1}$  & 79.0$_{\uparrow2.9}$  \\
      w/ GT, w/ LT & 66.4$_{\uparrow4.7}$  & 80.7$_{\uparrow4.6}$ \\
      \bottomrule
    \end{tabular}
    \caption{Ablation study on the proposed GLT module. GT and LT denote the global transformer block and local transformer block, respectively.}
    \label{table4}
\end{table}

\noindent\textbf{Sparse Sampling Size.} The sparse sampling size $m$ in global transformer block is an important hyper-parameter. A small size will be inefficient to capture global object prior, while using a large size will cause redundant information. Therefore, we conduct an experiment to evaluate the performance of using different sparse sampling size $m$. As shown in the top part of Table \ref{table5}, the best performance is obtained when $m=16$. Although there is a slight performance drop in the case of $m=32$, the number of floating point operations per second (FLOPs) becomes much larger.

\noindent\textbf{KNN Samlping Size.} In addition, we evaluate the performance of using different KNN sampling size $n$. The results are presented in the bottom part of Table \ref{table5}. When $n=8$, the local patch range is too small to effectively integrate information from surrounding points, thus limiting the performance. While in the case of $n=24$, more noisy information is included, thus degrading the feature representation of seed points. When $n=16$, it achieves the best performance of 68.2 and 82.1 in terms of \textit{Success} and \textit{Precision}, respectively. In this paper, we set $m=16$ and $n=16$ by default in all the experiments if not specified.

\begin{table}[!ht]
    \centering
    \begin{tabular}{ccccc}
      \toprule
      & Size & \textit{Success} & \textit{Precision} & FLOPs / G \\
      \midrule
      \midrule
      \multirow{3}{*}{SS ($m$)}
      & 4 & 61.3  & 75.9 & 1.14\\
      & 8 &  64.8 & 78.6 & 1.95\\
      & 16 & \textbf{68.2} & \textbf{82.1} & 3.56\\
      & 32 & 67.6 & 81.9 & 6.79\\
      \midrule
      \midrule
      \multirow{3}{*}{KS ($n$)}
      & 8 &  66.3 & 79.7 & 1.95\\
      & 16 & \textbf{68.2} & \textbf{82.1} & 3.56\\
      & 24 & 65.7 & 79.4 & 5.18\\
      \bottomrule
    \end{tabular}
    \caption{Ablation studies on the sparse sampling (SS) size and KNN sampling (KS) size.}
    \label{table5}
\end{table}

\subsection{Inference Speed} Similar to \cite{p2b,mlvsnet}, we calculate the tracking speed by counting the average running time of all frames on the Car category of KITTI dataset. GLT-T runs at 30 fps on a single NVIDIA 1080Ti GPU, including 8.7 ms for point cloud pre-processing, 24.1 ms for network forward computation and 0.6 ms for post-processing.

\section{Conclusion}
This work introduces a novel global-local transformer voting scheme for 3D object tracking (GLT-T) on point clouds. GLT-T promotes generating high quality 3D proposals to enhance performance. To learn strong representation for predicting offsets, we develop a cascaded global-local transformer module to integrate both object- and patch-aware prior. Moreover, a simple yet effective training strategy is designed, guiding the model to pay more attention to important seed points. Extensive experiments on KITTI and NuScenes benchmarks validate the effectiveness of GLT-T. We hope our voting scheme can serve as a basic component in future research to improve the tracking performance.
\noindent\textbf{Limitation and Discussion.} GLT-T follows the Siamese network-based appearance matching paradigm and therefore inherits its drawbacks, e.g., it is hard to deal with extremely sparse point clouds, where there is no sufficient information to perform a favorable appearance match and generate reliable seed points. One possible solution is to aggregate temporal contexts from previous multi-frame point clouds.



\begin{thebibliography}{42}
\providecommand{\natexlab}[1]{#1}

\bibitem[{Asvadi et~al.(2016)Asvadi, Girao, Peixoto, and Nunes}]{rgbd1}
Asvadi, A.; Girao, P.; Peixoto, P.; and Nunes, U. 2016.
\newblock 3D object tracking using RGB and LIDAR data.
\newblock In \emph{2016 IEEE 19th International Conference on Intelligent
  Transportation Systems (ITSC)}, 1255--1260. IEEE.

\bibitem[{Bibi, Zhang, and Ghanem(2016)}]{rgbd4}
Bibi, A.; Zhang, T.; and Ghanem, B. 2016.
\newblock 3d part-based sparse tracker with automatic synchronization and
  registration.
\newblock In \emph{Proceedings of the IEEE Conference on Computer Vision and
  Pattern Recognition}, 1439--1448.

\bibitem[{Caesar et~al.(2020)Caesar, Bankiti, Lang, Vora, Liong, Xu, Krishnan,
  Pan, Baldan, and Beijbom}]{nuscenes}
Caesar, H.; Bankiti, V.; Lang, A.~H.; Vora, S.; Liong, V.~E.; Xu, Q.; Krishnan,
  A.; Pan, Y.; Baldan, G.; and Beijbom, O. 2020.
\newblock nuscenes: A multimodal dataset for autonomous driving.
\newblock In \emph{Proceedings of the IEEE/CVF conference on computer vision
  and pattern recognition}, 11621--11631.

\bibitem[{Carion et~al.(2020)Carion, Massa, Synnaeve, Usunier, Kirillov, and
  Zagoruyko}]{detr}
Carion, N.; Massa, F.; Synnaeve, G.; Usunier, N.; Kirillov, A.; and Zagoruyko,
  S. 2020.
\newblock End-to-end object detection with transformers.
\newblock In \emph{European conference on computer vision}, 213--229. Springer.

\bibitem[{Cui et~al.(2021)Cui, Fang, Shan, Gu, and Zhou}]{lttr}
Cui, Y.; Fang, Z.; Shan, J.; Gu, Z.; and Zhou, S. 2021.
\newblock 3d object tracking with transformer.
\newblock arXiv:2110.14921.

\bibitem[{Fang et~al.(2020)Fang, Zhou, Cui, and Scherer}]{3d-siamrpn}
Fang, Z.; Zhou, S.; Cui, Y.; and Scherer, S. 2020.
\newblock {3d-siamrpn: An end-to-end learning method for real-time 3d single
  object tracking using raw point cloud}.
\newblock \emph{IEEE Sensors Journal}, 21(4): 1019--1026.

\bibitem[{Geiger, Lenz, and Urtasun(2012)}]{kitti}
Geiger, A.; Lenz, P.; and Urtasun, R. 2012.
\newblock Are we ready for autonomous driving? the kitti vision benchmark
  suite.
\newblock In \emph{2012 IEEE conference on computer vision and pattern
  recognition}, 3354--3361. IEEE.

\bibitem[{Giancola, Zarzar, and Ghanem(2019)}]{sc3d}
Giancola, S.; Zarzar, J.; and Ghanem, B. 2019.
\newblock {Leveraging Shape Completion for 3D Siamese Tracking}.
\newblock In \emph{IEEE/CVF Conference on Computer Vision and Pattern
  Recognition}, 1359--1368.

\bibitem[{Glorot, Bordes, and Bengio(2011)}]{relu}
Glorot, X.; Bordes, A.; and Bengio, Y. 2011.
\newblock Deep sparse rectifier neural networks.
\newblock In \emph{Proceedings of the fourteenth international conference on
  artificial intelligence and statistics}, 315--323. JMLR Workshop and
  Conference Proceedings.

\bibitem[{Guo et~al.(2021)Guo, Cai, Liu, Mu, Martin, and Hu}]{pct}
Guo, M.-H.; Cai, J.-X.; Liu, Z.-N.; Mu, T.-J.; Martin, R.~R.; and Hu, S.-M.
  2021.
\newblock Pct: Point cloud transformer.
\newblock \emph{Computational Visual Media}, 7(2): 187--199.

\bibitem[{Hui et~al.(2021)Hui, Wang, Cheng, Xie, and Yang}]{v2b}
Hui, L.; Wang, L.; Cheng, M.; Xie, J.; and Yang, J. 2021.
\newblock {3D Siamese Voxel-to-BEV Tracker for Sparse Point Clouds}.
\newblock In \emph{Advances in Neural Information Processing Systems},
  28714--28727.

\bibitem[{Kart et~al.(2019)Kart, Lukezic, Kristan, Kamarainen, and
  Matas}]{rgbd5}
Kart, U.; Lukezic, A.; Kristan, M.; Kamarainen, J.-K.; and Matas, J. 2019.
\newblock Object tracking by reconstruction with view-specific discriminative
  correlation filters.
\newblock In \emph{Proceedings of the IEEE/CVF Conference on Computer Vision
  and Pattern Recognition}, 1339--1348.

\bibitem[{Kristan et~al.(2021)Kristan, Matas, Leonardis, Felsberg, Pflugfelder,
  K{\"a}m{\"a}r{\"a}inen, Chang, Danelljan, Cehovin, Luke{\v{z}}i{\v{c}}
  et~al.}]{vot2021}
Kristan, M.; Matas, J.; Leonardis, A.; Felsberg, M.; Pflugfelder, R.;
  K{\"a}m{\"a}r{\"a}inen, J.-K.; Chang, H.~J.; Danelljan, M.; Cehovin, L.;
  Luke{\v{z}}i{\v{c}}, A.; et~al. 2021.
\newblock The ninth visual object tracking vot2021 challenge results.
\newblock In \emph{Proceedings of the IEEE/CVF International Conference on
  Computer Vision}, 2711--2738.

\bibitem[{Lan et~al.(2022)Lan, Zhang, He, and Zhang}]{lan}
Lan, M.; Zhang, J.; He, F.; and Zhang, L. 2022.
\newblock Siamese Network with Interactive Transformer for Video Object
  Segmentation.
\newblock In \emph{Proceedings of the AAAI Conference on Artificial
  Intelligence}, 1228--1236.

\bibitem[{Liu et~al.(2018)Liu, Jing, Nie, Gao, Liu, and Jiang}]{rgbd2}
Liu, Y.; Jing, X.-Y.; Nie, J.; Gao, H.; Liu, J.; and Jiang, G.-P. 2018.
\newblock Context-aware three-dimensional mean-shift with occlusion handling
  for robust object tracking in RGB-D videos.
\newblock \emph{IEEE Transactions on Multimedia}, 21(3): 664--677.

\bibitem[{Liu et~al.(2021)Liu, Lin, Cao, Hu, Wei, Zhang, Lin, and
  Guo}]{liu2021swin}
Liu, Z.; Lin, Y.; Cao, Y.; Hu, H.; Wei, Y.; Zhang, Z.; Lin, S.; and Guo, B.
  2021.
\newblock Swin transformer: Hierarchical vision transformer using shifted
  windows.
\newblock In \emph{Proceedings of the IEEE/CVF International Conference on
  Computer Vision}, 10012--10022.

\bibitem[{Lu et~al.(2022)Lu, Xie, Wei, Xu, and Li}]{ptsurvey}
Lu, D.; Xie, Q.; Wei, M.; Xu, L.; and Li, J. 2022.
\newblock Transformers in 3D Point Clouds: A Survey.
\newblock arxiv:2205.07417.

\bibitem[{Ma et~al.(2020)Ma, Tian, Xu, Huang, and Li}]{detection2}
Ma, W.; Tian, T.; Xu, H.; Huang, Y.; and Li, Z. 2020.
\newblock Aabo: Adaptive anchor box optimization for object detection via
  bayesian sub-sampling.
\newblock In \emph{European Conference on Computer Vision}, 560--575. Springer.

\bibitem[{Nie et~al.(2022{\natexlab{a}})Nie, He, Yang, Gao, and
  Dong}]{learning}
Nie, J.; He, Z.; Yang, Y.; Gao, M.; and Dong, Z. 2022{\natexlab{a}}.
\newblock Learning Localization-aware Target Confidence for Siamese Visual
  Tracking.
\newblock \emph{IEEE Transactions on Multimedia}.

\bibitem[{Nie et~al.(2022{\natexlab{b}})Nie, Wu, He, Gao, and Dong}]{spreading}
Nie, J.; Wu, H.; He, Z.; Gao, M.; and Dong, Z. 2022{\natexlab{b}}.
\newblock Spreading Fine-grained Prior Knowledge for Accurate Tracking.
\newblock \emph{IEEE Transactions on Circuits and Systems for Video
  Technology}.

\bibitem[{Pieropan et~al.(2015)Pieropan, Bergstr{\"o}m, Ishikawa, and
  Kjellstr{\"o}m}]{rgbd3}
Pieropan, A.; Bergstr{\"o}m, N.; Ishikawa, M.; and Kjellstr{\"o}m, H. 2015.
\newblock Robust 3d tracking of unknown objects.
\newblock In \emph{2015 IEEE International Conference on Robotics and
  Automation (ICRA)}, 2410--2417. IEEE.

\bibitem[{Qi et~al.(2019)Qi, Litany, He, and Guibas}]{votenet}
Qi, C.~R.; Litany, O.; He, K.; and Guibas, L.~J. 2019.
\newblock Deep hough voting for 3d object detection in point clouds.
\newblock In \emph{proceedings of the IEEE/CVF International Conference on
  Computer Vision}, 9277--9286.

\bibitem[{Qi et~al.(2017{\natexlab{a}})Qi, Su, Mo, and Guibas}]{pointnet}
Qi, C.~R.; Su, H.; Mo, K.; and Guibas, L.~J. 2017{\natexlab{a}}.
\newblock Pointnet: Deep learning on point sets for 3d classification and
  segmentation.
\newblock In \emph{Proceedings of the IEEE conference on computer vision and
  pattern recognition}, 652--660.

\bibitem[{Qi et~al.(2017{\natexlab{b}})Qi, Yi, Su, and Guibas}]{pointnet++}
Qi, C.~R.; Yi, L.; Su, H.; and Guibas, L.~J. 2017{\natexlab{b}}.
\newblock Pointnet++: Deep hierarchical feature learning on point sets in a
  metric space.
\newblock \emph{Advances in neural information processing systems}, 30.

\bibitem[{Qi et~al.(2020)Qi, Feng, Cao, Zhao, and Xiao}]{p2b}
Qi, H.; Feng, C.; Cao, Z.; Zhao, F.; and Xiao, Y. 2020.
\newblock {P2B: Point-to-Box Network for 3D Object Tracking in Point Clouds}.
\newblock In \emph{IEEE/CVF Conference on Computer Vision and Pattern
  Recognition}, 6328--6337.

\bibitem[{Shan et~al.(2021)Shan, Zhou, Fang, and Cui}]{ptt}
Shan, J.; Zhou, S.; Fang, Z.; and Cui, Y. 2021.
\newblock {PTT: Point-Track-Transformer Module for 3D Single Object Tracking in
  Point Clouds}.
\newblock In \emph{IEEE/RSJ International Conference on Intelligent Robots and
  Systems}, 1310--1316.

\bibitem[{Tian et~al.(2019)Tian, Shen, Chen, and He}]{fcos}
Tian, Z.; Shen, C.; Chen, H.; and He, T. 2019.
\newblock Fcos: Fully convolutional one-stage object detection.
\newblock In \emph{Proceedings of the IEEE/CVF international conference on
  computer vision}, 9627--9636.

\bibitem[{Vaswani et~al.(2017)Vaswani, Shazeer, Parmar, Uszkoreit, Jones,
  Gomez, Kaiser, and Polosukhin}]{aiayn}
Vaswani, A.; Shazeer, N.; Parmar, N.; Uszkoreit, J.; Jones, L.; Gomez, A.~N.;
  Kaiser, {\L}.; and Polosukhin, I. 2017.
\newblock Attention is all you need.
\newblock \emph{Advances in neural information processing systems}, 30.

\bibitem[{Wang et~al.(2021{\natexlab{a}})Wang, Cao, Zhang, and
  Tao}]{wang2021fp}
Wang, W.; Cao, Y.; Zhang, J.; and Tao, D. 2021{\natexlab{a}}.
\newblock Fp-detr: Detection transformer advanced by fully pre-training.
\newblock In \emph{International Conference on Learning Representations}.

\bibitem[{Wang et~al.(2022)Wang, Zhang, Cao, Shen, and Tao}]{wang2022towards}
Wang, W.; Zhang, J.; Cao, Y.; Shen, Y.; and Tao, D. 2022.
\newblock Towards Data-Efficient Detection Transformers.
\newblock In \emph{European conference on computer vision}.

\bibitem[{Wang et~al.(2021{\natexlab{b}})Wang, Xie, Lai, Wu, Long, and
  Wang}]{mlvsnet}
Wang, Z.; Xie, Q.; Lai, Y.-K.; Wu, J.; Long, K.; and Wang, J.
  2021{\natexlab{b}}.
\newblock {MLVSNet: Multi-level Voting Siamese Network for 3D Visual Tracking}.
\newblock In \emph{IEEE/CVF International Conference on Computer Vision},
  3081--3090.

\bibitem[{Wu, Lim, and Yang(2013)}]{otb}
Wu, Y.; Lim, J.; and Yang, M.-H. 2013.
\newblock Online object tracking: A benchmark.
\newblock In \emph{Proceedings of the IEEE conference on computer vision and
  pattern recognition}, 2411--2418.

\bibitem[{Xu et~al.(2022)Xu, Zhang, Zhang, and Tao}]{hpe}
Xu, Y.; Zhang, J.; Zhang, Q.; and Tao, D. 2022.
\newblock ViTPose: Simple Vision Transformer Baselines for Human Pose
  Estimation.
\newblock arxiv:2204.12484.

\bibitem[{Xu et~al.(2021)Xu, Zhang, Zhang, and Tao}]{xu2021vitae}
Xu, Y.; Zhang, Q.; Zhang, J.; and Tao, D. 2021.
\newblock Vitae: Vision transformer advanced by exploring intrinsic inductive
  bias.
\newblock \emph{Advances in Neural Information Processing Systems}, 34:
  28522--28535.

\bibitem[{Zhang and Tao(2020)}]{things}
Zhang, J.; and Tao, D. 2020.
\newblock {Empowering things with intelligence: a survey of the progress,
  challenges, and opportunities in artificial intelligence of things}.
\newblock \emph{IEEE Internet of Things Journal}, 8(10): 7789--7817.

\bibitem[{Zhang et~al.(2022{\natexlab{a}})Zhang, Xu, Zhang, and
  Tao}]{zhang2022vitaev2}
Zhang, Q.; Xu, Y.; Zhang, J.; and Tao, D. 2022{\natexlab{a}}.
\newblock Vitaev2: Vision transformer advanced by exploring inductive bias for
  image recognition and beyond.
\newblock \emph{arXiv preprint arXiv:2202.10108}.

\bibitem[{Zhang et~al.(2022{\natexlab{b}})Zhang, Xu, Zhang, and
  Tao}]{zhang2022vsa}
Zhang, Q.; Xu, Y.; Zhang, J.; and Tao, D. 2022{\natexlab{b}}.
\newblock VSA: Learning Varied-Size Window Attention in Vision Transformers.
\newblock In \emph{European conference on computer vision}.

\bibitem[{Zhang et~al.(2020)Zhang, Chi, Yao, Lei, and Li}]{detection1}
Zhang, S.; Chi, C.; Yao, Y.; Lei, Z.; and Li, S.~Z. 2020.
\newblock Bridging the gap between anchor-based and anchor-free detection via
  adaptive training sample selection.
\newblock In \emph{Proceedings of the IEEE/CVF conference on computer vision
  and pattern recognition}, 9759--9768.

\bibitem[{Zhao et~al.(2021)Zhao, Jiang, Jia, Torr, and Koltun}]{pt}
Zhao, H.; Jiang, L.; Jia, J.; Torr, P.~H.; and Koltun, V. 2021.
\newblock Point transformer.
\newblock In \emph{Proceedings of the IEEE/CVF International Conference on
  Computer Vision}, 16259--16268.

\bibitem[{Zheng et~al.(2021)Zheng, Yan, Gao, Zhao, Zhang, Li, and Cui}]{bat}
Zheng, C.; Yan, X.; Gao, J.; Zhao, W.; Zhang, W.; Li, Z.; and Cui, S. 2021.
\newblock {Box-Aware Feature Enhancement for Single Object Tracking on Point
  Clouds}.
\newblock In \emph{IEEE/CVF International Conference on Computer Vision},
  13179--13188.

\bibitem[{Zhou et~al.(2022)Zhou, Luo, Luo, Liu, Pan, Cai, Zhao, and Lu}]{pttr}
Zhou, C.; Luo, Z.; Luo, Y.; Liu, T.; Pan, L.; Cai, Z.; Zhao, H.; and Lu, S.
  2022.
\newblock {PTTR: Relational 3D Point Cloud Object Tracking with Transformer}.
\newblock In \emph{IEEE/CVF Conference on Computer Vision and Pattern
  Recognition}, 8531--8540.

\bibitem[{Zou et~al.(2020)Zou, Cui, Kong, Zhang, Liu, Wen, and Li}]{f-siamese}
Zou, H.; Cui, J.; Kong, X.; Zhang, C.; Liu, Y.; Wen, F.; and Li, W. 2020.
\newblock {F-Siamese Tracker: A Frustum-based Double Siamese Network for 3D
  Single Object Tracking}.
\newblock In \emph{IEEE/RSJ International Conference on Intelligent Robots and
  Systems}, 8133--8139.

\end{thebibliography}

\end{document}